# Time Series Prediction by Multi-task GPR with Spatiotemporal Information Transformation

Peng Tao#, Xiaohu Hao# , Jie Cheng and Luonan Chen*

*Abstract*—**Making an accurate prediction of an unknown system only from a short-term time series is difficult due to the lack of sufficient information, especially in a multi-step-ahead manner. However, a high-dimensional short-term time series contains rich dynamical information, and also becomes increasingly available in many fields. In this work, by exploiting spatiotemporal information (STI) transformation scheme that transforms such high-dimensional/spatial information to temporal information, we developed a new method called MT-GPRMachine to achieve accurate prediction from a short-term time series. Specifically, we first construct a specific multi-task GPR which is multiple linked STI mappings to transform high-dimensional/spatial information into temporal/dynamical information of any given target variable, and then makes multi-step-ahead prediction of the target variable by solving those STI mappings. The multi-step-ahead prediction results on various synthetic and real-world datasets clearly validated that MT-GPRMachine outperformed other existing approaches.**

## I. INTRODUCTION

HOW to make an accurate multi-step-ahead prediction based on only the observed time-series data of a nonlinear dynamical system, especially short-term data, has been attracting more and more attention in many fields such as computational biology, ecology, geoscience and econometrics[1-6]. However, multi-step-ahead prediction is a difficult and challenging task due to the complicated behavior of the dynamical system and the lack of sufficient information or data[7-9]. Here we first summarize the existing methods of time series prediction into serval categories: classical statistical regression-based methods, currently popular machine learning-based methods, and embedding theorem-based methods, and then introduce our new method, MT-GPRMachine.

Existing statistical regression-based methods including AR and ARIMA[10, 11], robust regression[12], exponential smoothing[13, 14], Lasso and AdaLasso[15], etc. Most of those methods mainly use the data of a single or target variable itself, and the abundant spatial information of other observed variables in the time series are all unexploited. Nevertheless, the observed high-dimensional data contain rich information, which can be explored to compensate the short time series for making a more accurate prediction. Besides, most of those methods are for one-step-ahead prediction, and thus inevitably, errors from the one-step-ahead prediction are accumulated along with the time steps, making the statistical regression-based methods generally perform poorly in multi-step-ahead prediction[7, 16].

Gaussian process regression (GPR)[17-19] is another category of statistical regression methods. The Gaussian process generates data located throughout some domain such that any finite subset of the range follows a multivariate Gaussian distribution. With a number of observed samples available, GPR establishes the distribution of those data and further makes predictions for the target variable with associate observations given. Based on the basic GPR, serval derived GPR methods have been proposed and successfully applied in various fields, such as GP state-space[20, 21], warped/manifold GPs[22, 23], and multi-task learning GP[24-27]. Generally, most of the GP-based regression methods require the part/whole of the future observations to estimate/calibrate the system dynamics/state, which becomes an obstacle in making predictions of the system dynamics based on only the previous observed time series. Especially for the case that the future state of the system is difficult/consumptive to be observed.

Manuscript received on 18 Jan 2022. This work is supported by National Key R&D Program of China (No. 2017YFA0505500), by the Strategic Priority Project of CAS (No. XDB38000000), by the Natural Science Foundation of China (Nos. 31771476, 31930022 and 12026608), by Shanghai Municipal Science and Technology Major Project (No. 2017SHZDZX01), and by the Japan Science and Technology Agency Moonshot R&D (No. JPMJMS2021) (*Corresponding author: Luonan Chen; #These authors contributed equally to this work).

P. Tao is with Key Laboratory of Systems Health Science of Zhejiang Province, Hangzhou Institute for Advanced Study, University of Chinese Academy of Sciences, Hangzhou 310024 China; State Key Laboratory of Cell Biology, Shanghai Institute of Biochemistry and Cell Biology, Center for Excellence in Molecular Cell Science, Chinese Academy of Sciences, Shanghai 200031 China (taopeng@ucas.ac.cn).

X. Hao is with State Key Laboratory of Cell Biology, Shanghai Institute of Biochemistry and Cell Biology, Center for Excellence in Molecular Cell Science, Chinese Academy of Sciences, Shanghai 200031 China (haoxiaohu@sibcb.ac.cn).

J. Cheng is with HUAWEI Technologies Co., Ltd, Shenzhen 518129, China (chengjie8@huawei.com).

L. Chen is with Key Laboratory of Systems Health Science of Zhejiang Province, Hangzhou Institute for Advanced Study, University of Chinese Academy of Sciences, Hangzhou 310024 China; State Key Laboratory of Cell Biology, Shanghai Institute of Biochemistry and Cell Biology, Center for Excellence in Molecular Cell Science, Chinese Academy of Sciences, Shanghai 200031 China; and School of Life Science and Technology, ShanghaiTech University, Shanghai 201210, China (lnchen@sibcb.ac.cn).

Machine learning-based methods such as long-short-term-memory (LSTM) neural networks [28, 29], reservoir computing[30, 31] and autoreservoir computing[16] are widely used in time series prediction and other aspects in many fields. While one of the biggest drawbacks for machine learning-based methods is that the network should be trained by a large number of samples for learning the dynamic characteristics of a system to avoid the overfitting problem. High-dimensionality of the real-world data may also encumber the computational efficiency of the training process. Those defects make the general machine learning-based methods difficult to be applied for time series analysis with only short-term data available. In addition, from a theoretical viewpoint, inexplicable black-box models of the neural network-based methods are still an obstacle for understanding the prediction mechanism.

Embedding theory[32-34] provides a specific linkage between state space and temporal space of a dynamical system (the dynamical behavior of a dynamical system can be reconstructed from the lags of a single variable[35]) and thus widely used in time series analysis. State space reconstruction (SSR)[36], multi-view embedding (MVE)[37], randomly distributed embedding (RDE)[38], attractor ranked radial basis function network (AR-RBFN)[39] and anticipated learning machine (ALM)[40] are representative embedding theorem-based methods. Most of those methods are designed for one-step-ahead prediction which makes their application limited for multi-step-ahead prediction.

Accordingly, obtaining accurate multi-step-ahead prediction rather than multiple one-step-ahead predictions is demanded in real-world applications and urgent to be solved. Also, even if there are long-term data, the prediction based on short time series is also practical and effective because many complex systems are so highly time-varying that only their recent short-term data are informative for future prediction.

Aiming at the multi-step-ahead prediction problem with short-term but high-dimensional time-series data, we proposed MT-GPRMachine, a multi-task learning-based Gaussian process regression method for addressing the aforementioned challenges. Specifically, the spatiotemporal information (STI) transformation matrix was designed to bridge the temporal evolution (multiple time steps) of a single variable and spatial information (one time step) of multiple variables by sequentially stacking the multiple one-step-delay STI[38] equations into a matrix. By such stacking operation, the original small sample size can be virtually expanded and thus alleviating the problem of small sample size or short time series. In the formed STI matrix, each row represents a nonlinear mapping with a certain number of samples $\mathbf{X}$ and corresponding target values $\mathbf{Y}$, which can be solved by GPR. In addition, the special Hankel matrix[41] structure of the formed STI matrix provides a linkage between the mappings, which can be considered as constraints and fully exploited by multi-task GPR, so that we can solve the linked mappings in the STI matrix conveniently and effectively. The multiple linked GPRs not only learn the mappings from the observations, but also learn them from each other, and actually, the cooperation among the linked GPRs helps to reduce the prediction error of each other. Once the mappings in the STI matrix have been solved, the multi-step-ahead prediction can be obtained accordingly. In other words, the major contribution of this work is that we design a specific multi-task GPR that collectively solves the nonlinear STI mappings, and thus provides accurate multi-step-ahead prediction in an efficient and effective manner.

Notice that different from the general multi-task GPR which simulates different targets from different tasks, MT-GPRMachine simulates the same target. In particular, MT-GPRMachine is constrained by the special STI matrix or Hankel matrix. Besides, different from RDE, MT-GPRMachine extends the one-step-ahead to the multi-step-ahead prediction by integrating all mappings in the STI matrix. Actually, MT-GPRMachine exploits the special Hankel matrix structure in the whole STI matrix, which can also be viewed as an integrated multi-task GPR. What is more, MT-GPRMachine is a nonparametric method with training and predicting processes conducted simultaneously by solving the linked mappings, which means that we have no information from the future state of the system and make predictions purely from the previously observed samples. Last but not least, MT-GPRMachine makes a prediction with an explicable mechanism because of the theoretical foundation of GPR. Our framework is shown in **Fig. 1**.

## II. METHODS

### A. STI Transformation Equation

Generally, there is no effective way to predict the short-term time series because most of the existing approaches require sufficiently long-term time-series data. For such a problem, the delay embedding theorem[34] provides a new way for connecting the future evolution of a target variable and the observed spatial information, and shows that the dynamical behavior or attractor of a system can be reconstructed from the lags of a single variable[32, 33, 35].

Based on the delay embedding theorem, the spatiotemporal information (STI) transformation equation has been proposed for bridging the high-dimensional spatial information and the temporal information of a target variable[38, 40]. Specifically, the STI equation can be written as

$$\mathbf{\Psi}(\mathbf{X}(t)) = \mathbf{Y}(t), \qquad (1)$$

and the corresponding matrix form is

$$
\begin{pmatrix}
\mathbf{\Psi}_1(\mathbf{X}(t_1)) & \mathbf{\Psi}_1(\mathbf{X}(t_2)) & \cdots & \mathbf{\Psi}_1(\mathbf{X}(t_M)) \\
\mathbf{\Psi}_2(\mathbf{X}(t_1)) & \mathbf{\Psi}_2(\mathbf{X}(t_2)) & \cdots & \mathbf{\Psi}_2(\mathbf{X}(t_M)) \\
\cdots & \cdots & \ddots & \cdots \\
\mathbf{\Psi}_L(\mathbf{X}(t_1)) & \mathbf{\Psi}_L(\mathbf{X}(t_2)) & \cdots & \mathbf{\Psi}_L(\mathbf{X}(t_M))
\end{pmatrix}
$$
$$
=
\begin{pmatrix}
y(t_1) & y(t_2) & \cdots & y(t_M) \\
y(t_2) & y(t_3) & \cdots & y(t_{M+1}) \\
\cdots & \cdots & \ddots & \cdots \\
y(t_L) & y(t_{L+1}) & \cdots & y(t_{M+L-1})
\end{pmatrix},
\qquad (2)
$$

where $\mathbf{X}(t_1)$ to $\mathbf{X}(t_M)$ represent $M$ observed system states (spatial information) at time points $t_1$ to $t_M$, which are $N$-dimension vectors containing $N$ variables $x_1(t)$ to $x_N(t)$, and $L$ is the delay embedding dimension. In general, $\mathbf{X}(t) = [x_1(t), x_2(t), \dots,$ $x_n(t), \dots, x_N(t)$]; $x_n(t)$ is the chosen target variable (temporal information) to be predicted, which is marked as $y(t)$ in **Eq.(2)**, i.e. $y(t)=x_n(t)$; $\Psi_1$ to $\Psi_L$ are the nonlinear mappings between $\mathbf{X}(t)$ and $y(t)$, where the condition $L>2d$ (where $d$ is a small positive number represents the box-counting dimension of an attractor of the dynamical system) should be satisfied[34].

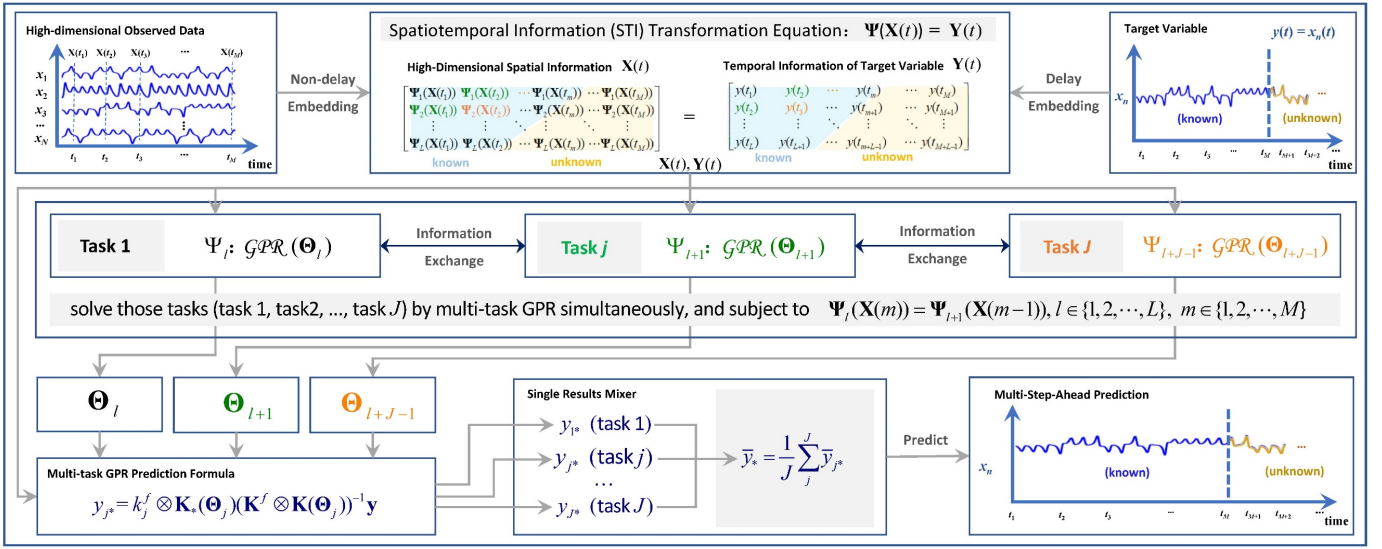

**Fig. 1.** Framework of MT-GPRMachine. MT-GPRMachine starts from the short-term but high-dimensional observed data, it bridges the spatial and temporal information through STI matrix. The correlated $J$ GPRs are simultaneously solved as $J$ tasks, with the information exchange and the constraints between all $J$ tasks, the optimized parameters can be obtained. Estimated values of the target variable from every GPRs can be then obtained for further combination for final prediction.

The $L-1$ future values of $y(t)$ after $t_M$ are included in $\mathbf{Y}(t)$. Accordingly, with all $\mathbf{X}(t)$ known, $L-1$ steps predictions can be made by solving the mappings in the STI matrix. In general, the maximal number of $L-1=M-1$ steps prediction can be made. The $L-1$ mappings are actually linked together due to the Hankel matrix form, specifically, $\Psi_{l-1}(\mathbf{X}(t_{m+1})) \equiv \Psi_l(\mathbf{X}(t_m)) \equiv \Psi_{l+1}(\mathbf{X}(t_{m-1}))$ always holds for $1<l<L$, $1<m\le M$, which can be fully exploited by GPR with the multi-task learning scheme introduced.

### B. Gaussian Process Regression

With the observed variables available, Gaussian process[17] is a powerful method to establish the distribution of those data. Further, with extra future observations, Gaussian process regression[18] can estimate the value of other target variables on the basis of the Gaussian process. The Gaussian process generates data located throughout some domain such that any finite subset of the range follows a multivariate Gaussian distribution. In any dataset $\mathbb{X} = \{\mathbf{X}(t_1), \mathbf{X}(t_2), \cdots, \mathbf{X}(t_M)\}$ the $M$ observations can be imagined as $M$ points sampled from some multivariate ($n$-variate) Gaussian distribution. Rather than claiming that $\Psi(\mathbf{X}(t))$ is related to a specific model (e.g., $\Psi(\mathbf{X}(t)) = a\mathbf{X}(t)+b$), a Gaussian process can represent $\Psi(\mathbf{X}(t))$ obliquely, but rigorously, by letting the data 'speak' more clearly for themselves. The Radial Basis Function (RBF) covariance function $k(\cdot) = k(\mathbf{X}(t), \mathbf{X}'(t))$ which is for distinguishing two observations, can be chosen as the kernel function

$$k(\cdot) = \sigma_f^2 \cdot \exp\left(\frac{-(\mathbf{X}(t)-\mathbf{X}'(t))^2}{2l^2}\right) + \sigma_n^2 \cdot \delta(\mathbf{X}(t)-\mathbf{X}'(t)), \quad (3)$$

where $\sigma_f^2$, $\sigma_n^2$, $l$ and $\delta$ are the signal variance, Gaussian noise, scaling factor of each variable in $\mathbf{X}(t)$ and Kronecker delta function, respectively.

Denoting any single point, e.g., $x_n(t_{M+1})$ to be the target variable which is to be predicted, marked as $y*$, it can be represented as a sample from a multivariate Gaussian distribution as

$$\begin{bmatrix} \mathbf{y} \\ y_* \end{bmatrix} \sim N\left(0, \begin{bmatrix} \mathbf{K} & \mathbf{K}_*^{\mathsf{T}} \\ \mathbf{K}_* & \mathbf{K}_{**} \end{bmatrix}\right), \quad (4)$$

where $\mathbf{K}$ is the covariance matrix of $M$ observed samples calculated according to **Eq. (3)**, $\mathbf{y}$ is the vector of observed target variables. Then, the probability of $y*$ follows

$$y_* \mid \mathbf{y} \sim N(\mathbf{K}_* \mathbf{K}^{-1} \mathbf{y}, \mathbf{K}_{**} - \mathbf{K}_* \mathbf{K}^{-1} \mathbf{K}_*^{\mathsf{T}}), \quad (5)$$

where $\mathbf{K}_* \mathbf{K}^{-1} \mathbf{y}$ is the best estimation of $y*$, marked as $\bar{y}_*$, $\mathbf{K}_{**} - \mathbf{K}_* \mathbf{K}^{-1} \mathbf{K}_*^{\mathsf{T}}$ refers to the uncertainty of the estimation, marked as $\text{var}(y_*)$, and the regression can be made accordingly, where $\mathbf{K}_* = k(\mathbf{X}_*(t), \mathbb{X})$, $\mathbf{K}_{**} = k(\mathbf{X}_*(t), \mathbf{X}_*(t))$, and $\mathbf{X}_*(t)$ is the assumed observation of the target point to be predicted. The reliability of regression depends on the selection of covariance function. Specifically, a set of parameters $\Theta = \{\sigma_f^2, \sigma_n^2, l\}$ need to be optimized by maximizing a posteriori estimate of $\Theta$ when $p(\Theta \mid \mathbb{X}, \mathbf{y})$ is at its greatest, which is equal to maximize the log marginal likelihood according to Bayes' theorem[42]

$$\log p(\mathbf{y} \mid \mathbb{X}, \Theta) = -\frac{1}{2} \mathbf{y}^{\mathsf{T}} \mathbf{K}^{-1} \mathbf{y} - \frac{1}{2} \log |\mathbf{K}| - \frac{M}{2} \log 2\pi. \quad (6)$$

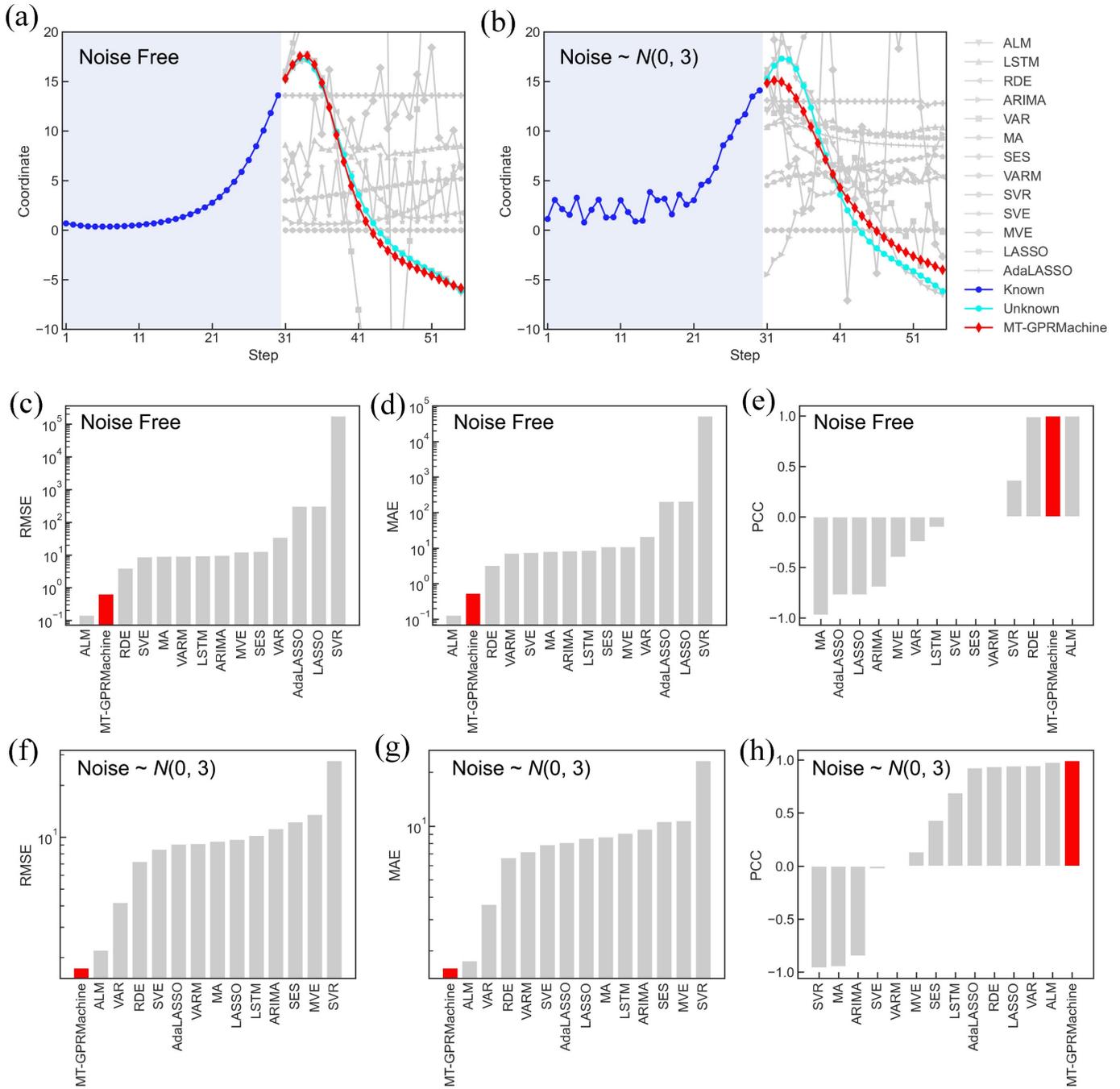

**Fig. 2.** Test results on a 90-dimension time-variant coupled Lorentz system (target variable is x_16). (**a** and **b**) are the predicted results for the noise-free and noise-polluted (Gaussian noise ~ $N(0, 3)$) cases, respectively. The blue, cyan and red lines represent the known (training, 30 time points), unknown (test, 25 time points) and predicted values (MT-GPRMachine) of the target variable, respectively. For the convenience of visualization, the predicted values of the other 13 methods are displayed in gray. The corresponding prediction performances, including MAE, RMSE and PCC, are shown in (**c-e**) for noise-free case, (**f-h**) for noise-polluted case, respectively.

## C. Multi-task Gaussian Process Regression

Here we briefly introduce the multi-task GPR[24, 25], which actually consider similar learning tasks together by introducing an extra kernel for measuring the difference between different tasks for getting benefit from each other. Specifically, the mathematical expressions of multi-task GPR are similar to single-task GPR, while the main difference is about the kernel, the kernel $\boldsymbol{k}(\cdot)$ used in the multi-task GPR is expressed as

$$\boldsymbol{k}(\cdot) = \mathbf{K}^f k(\cdot), \qquad (7)$$

where $\mathbf{K}^f$ is semi-definite matrix (covariance matrix between tasks) which specifies the correlation between different tasks, $k(\cdot)$ is the covariance function over inputs.

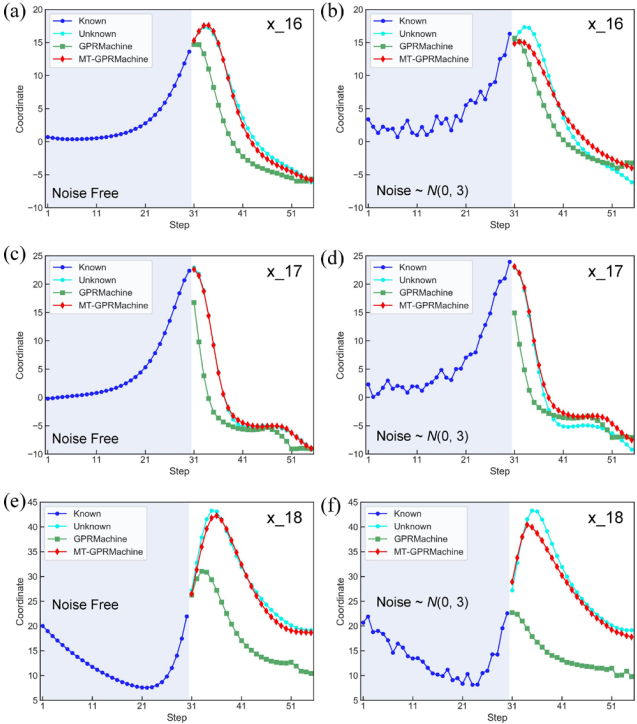

**Fig. 3.** The effects brought by introducing the multi-task learning scheme into GPR. The predicted results under the noise-free case for targets x_16, x_17 and x_18 are shown in (**a**, **c** and **e**). The corresponding results under the noise polluted (Gaussian noise $\sim N(0, 3)$) case are shown in (**b**, **d** and **f**). The blue and cyan lines represent the known (training, 30 time points) and unknown (test, 25 time points) values of the target variable, respectively. The predicted values of GPRMachine ($J$=1) and MT-GPRMachine ($J$=5) are marked by green squares and red diamonds, respectively.

Similar to GPR, the multi-task GPR can be solved by maximizing a posteriori estimate of $\mathbf{\Theta}$ and $\mathbf{K}^f$ when $p(\mathbf{\Theta}, \mathbf{K}^f \mid \mathbb{X}, \mathbf{y})$ is at its greatest, which is equal to maximize the log marginal likelihood

$$\log p(\mathbf{\Theta}, \mathbf{K}^f \mid \mathbb{X}, \mathbf{y}), \quad (8)$$

where $\mathbf{K}^f$ is the semi-definite matrix contains parameters to be optimized. The detailed expression of **Eq. (8)** can be found in reference[24]. Then the estimation of a target point to be predicted by task $j$, marked as $\overline{y}_{j*}$, can be expressed as

$$\overline{y}_{j*} = k_j^f \otimes \mathbf{K}_*(\mathbf{K}^f \otimes \mathbf{K} + \sigma_n^2 \mathbf{I})^{-1} \mathbf{y}, \quad (9)$$

where $\mathbf{I}$ is the unit matrix, $k_j^f$ refers to the $j$-th task which is the $j$-th column of $\mathbf{K}^f$, $\otimes$ denotes the Kronecker product.

### D. Solving the STI Mappings by Multi-task GPR

Simply, once the STI matrix has been established, the $(l-1)$-step-ahead prediction can be made by solving the single mapping $\mathbf{\Psi}_l$. Taking the $l$-th mapping under consideration, there are $M-l+1$ known target values $y(t_1)$ to $y(t_M)$ in total, and other $l-1$ target values $y(t_{M+1})$ to $y(t_{M+l-1})$ are the future values of the target variable to be predicted. The original observed time series data are separated into two parts: the observations with previously known target values which are used for learning the parameters $\mathbf{\Theta}$, and the observations with future unknown target values which can be regarded as the inputs of the trained GPR model to predict the future target values. That makes the multi-step-ahead prediction from only the previously observed short-term time series possible and thus different from other methods.

While the predictions made in such a way are not accurate due to the insufficient training samples, especially for the last mapping, e.g., $\mathbf{\Psi}_L$ ($L$=$M-1$) with only one training sample, which probably failed in making a prediction. Notably, the Hankel matrix structure of STI matrix lets the mappings organically linked together. More specific, aiming at predicting target variable $x_n$, as shown in **Eq. (2)** and **Fig. 1**, mappings $\mathbf{\Psi}_1$ to $\mathbf{\Psi}_L$ share $M-l+1$ observed samples $\mathbf{X}(t_1)$ to $\mathbf{X}(t_{M-l+1})$, and their corresponding target values $x_n(t_1)$ (or $y(t_1)$) to $x_n(t_M)$ (or $y(t_M)$). Naturally, the highly related mappings in the STI matrix and its special structure can be fully exploited by the multi-task learning GPR.

By introducing an extra kernel $\mathbf{K}^f$ which specifies inter-task similarities, the $J$ neighboring mappings in the STI matrix can be solved together by maximizing the posterior estimate of $\mathbf{\Theta}$ and $\mathbf{K}^f$ when $p(\mathbf{\Theta}, \mathbf{K}^f \mid \mathbb{X}, \mathbf{y})$ is at its greatest, which is equal to maximize the log marginal likelihood as expressed in **Eq. (8)**. Different from general multi-task learning GPR, all $J$ tasks in our method learn the same target. And the solving process is further constrained by the Hankel matrix structure of the sub-STI matrix (which contains $J$ rows of the original STI matrix) automatically due to the shared training samples of those mappings.

Taking use of the off-the-shelf optimization method integrated in the GPflow software, MT-GPRMachine orderly learns the parameters $\mathbf{\Theta}$ and $\mathbf{K}^f$ for every $J$ continuous mappings with the previous part of observations with known target values. The predictions of the future target values are given by the $J$ trained GPRs with the later part of observations as inputs. With all mappings trained, the predictions for the target variable at $L-1$ time points have been made as well. Then, the average value of the prediction at every time point is taken as the final prediction. The overall solving process is summarized as pseudo code in the supporting information, **SI**.

## III. EXPERIMENTS

### A. Datasets

MT-GPRMachine was first tested on two synthetic datasets: a 90-dimension time-variant coupled Lorentz system[43] and a 64-dimension time-variant coupled pendulum system[44]. Then, MT-GPRMachine was further applied on various real-world datasets: the typhoon eye trace (typhoon Marcus, 2018), plankton dataset (Northern Gulf, Mexico, 2003-2010), ground ozone level dataset (Houston, Galveston and Brazoria area, 1998-2004) and wireless channel dataset (HUAWEI Technologies Co., Ltd). How we generate the synthetic datasets and the detailed descriptions of the real-world datasets are listed in **SI**.

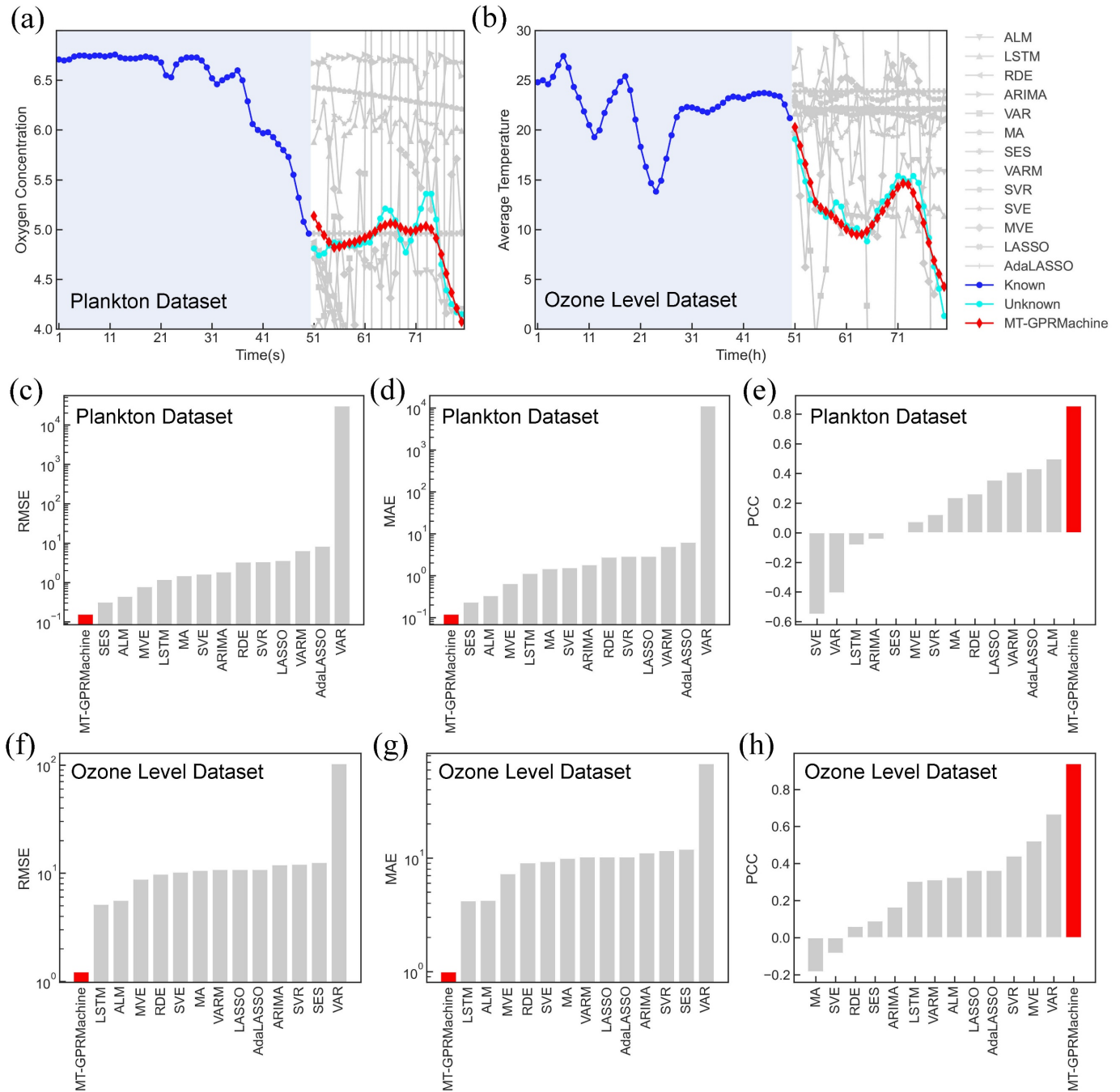

**Fig. 4.** Test the effectiveness of MT-GPRMachine on the (**a**) plankton dataset and (**b**) ozone level dataset. The blue, cyan and red lines represent the known (training, 50 time points), unknown (test, 30 time points) and predicted values of the target variables (oxygen concentration for plankton dataset, average temperature for ozone level dataset), respectively. For the convenience of visualization, the predicted values of the other 13 methods are displayed in gray. The corresponding prediction performances, including MAE, RMSE and PCC, are shown in (**c-e**) for the plankton dataset, (**f-h**) for the ozone level dataset, respectively.

### B. Comparison with Other Baselines

Based on the GPflow[45], scikit-learn[46] and other general Python packages, MT-GPRMachine was implemented with Python and run on an Intel Core i7 machine with 48 GB RAM, 2.40 GHz CPU, and 64-bit Ubuntu system. The parameter $J$ was set to 5 under consideration of the trade-off between prediction accuracy and computational complexity (computing speed). $L$ was set differently according to the observation length of the datasets. The settings of $B_l$, $B_u$ and $max\_run$ please refer to the public released source code.

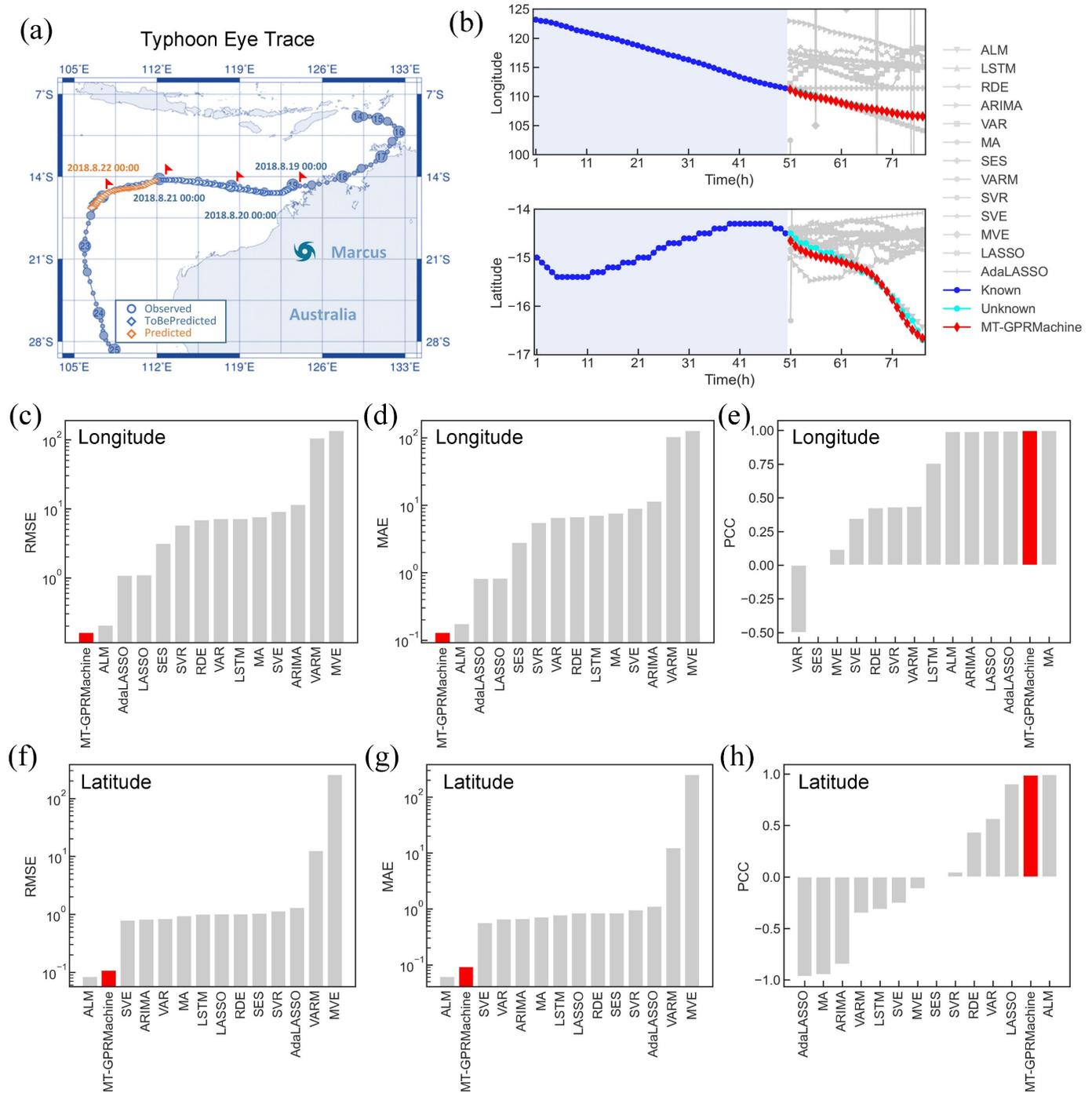

**Fig. 5.** Test the effectiveness of MT-GPRMachine on the typhoon eye trace dataset. (**a**) Comparison between the observed and predicted trace of eye center of the typhoon Marcus on the map. (**b**) The blue, cyan and red lines represent the known (training, 50 time points), unknown (test, 27 time points) and predicted values of the target variables (up: longitude, down: latitude), respectively. For the convenience of visualization, the predicted values of the other 13 methods are displayed in gray. The corresponding prediction performances, including MAE, RMSE and PCC, are shown in (**c-e**) for the longitude, (**f-h**) for the latitude, respectively.

MT-GPRMachine was compared with other 13 wildly used baselines including traditional statistical regression methods like ARIMA[10, 11], VAR[47], MA, SES[14] and VARM[48]; Machine learning methods like SVR[49] and LSTM[28, 29]; Embedding theory-based methods like SVE[50], MVE[37], RDE[38] and ALM[40]; Other methods like Lasso and AdaLasso[15]. Mean absolute error (MAE), root mean square error (RMSE), and Pearson correlation coefficient (PCC) were used for evaluating the prediction performance. MAE and RMSE reflect the numerical errors between the original and predicted values, and PCC shows the consistency of overall trends between the predicted and the original time series.

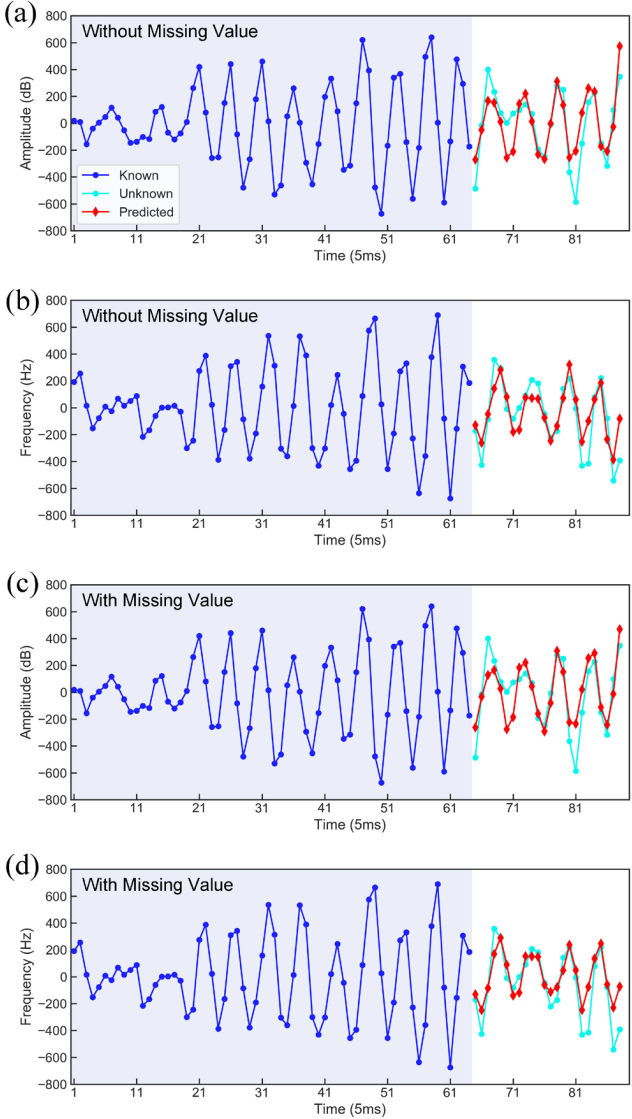

**Fig. 6.** Test the effectiveness of MT-GPRMachine on the wireless channel dataset. The predicted results for the amplitude (real part) and frequency (imaginary part) are shown in **a** and **b**, respectively. The corresponding predicted results for the training data with 10% missing values are shown in **c** and **d**, respectively. The blue, cyan and red lines represent the known (training, 64 time points), unknown (test, 24 time points) and predicted (our method) values of the corresponding target variable (3rd of 192 variables), respectively.

## IV. RESULTS

### A. Test on the Synthetic Datasets

To validate the ability of MT-GPRMachine in making the multi-step-ahead prediction with a short-term time series, a 90-dimension time-variant coupled Lorentz system was employed. Three target variables marked as x_16, x_17 and x_18 were chosen as targets, respectively. We made $L$-1=25 steps ahead prediction from a segment that includes 30 observed samples for the Lorentz system. Here we take x_16 as an example. As shown in **Fig. 2a**, when there is no noise, the prediction curve of MT-GPRMachine ('$n\_task$=5' or $J$=5) is consistent with the real curve. Compared with the other 13 benchmark algorithms (detailed results are listed in **Table S1**), its prediction performance (including MAE, RMSE and PCC) is second only to ALM (**Fig. 2c-e**). When there is noise pollution (Gaussian noise ~ $N(0, 3)$) in the observed data, the prediction performance of MT-GPRMachine exceeds that of ALM, and its three performance indicators are the highest among all methods. This result is also held for x_17 and x_18 (see **Fig. S1** and **S2**). These results show that MT-GPRMachine not only has excellent prediction performance, but also has strong noise immunity. Considering that real datasets often contain noise, MT-GPRMachine is more suitable for real datasets than ALM, which can also be seen from the practical application later.

We next analyze the effects brought by introducing the multi-task learning scheme into GPR. Also take the results on the Lorentz system as an example, as intuitively shown in **Fig. 3**, no matter whether or not there is noise pollution in the observation data (blue lines), the predicted values of MT-GPRMachine ($J$=5, red lines) are much closer to the original true values (cyan lines) than that of general GPRMachine ($J$=1, green lines, all other parameters were set equally, detailed results are shown in **Fig. S3** and **S4**). Quantitatively, for noise-free case, the performance of MT-GPRMachine was improved by 9.5% in terms of PCC, and in terms of MAE and RMSE, the indexes were decreased by 89.8% and 90.2%, respectively (see **Table S2** for more details). The result demonstrates that with the multi-task learning scheme introduced, the prediction accuracy was significantly improved. Another advantage of introducing the multi-task learning scheme is the increased stability of predicted results. After conducting the independent running of GPRMachine and MT-GPRMachine 30 times respectively, the average and standard deviation of MAE, RMSE and PCC are compared (**Table S2**), which shows that MT-GPRMahcine performed steadily with lower variance (decreased by 6.6%). To verify that the prediction performance of MT-GPRMachine is robust to different datasets, a 64-dimension coupled pendulum system was employed and similar results were obtained (see **Fig. S5** and **S6**).

### B. Application on real-world datasets

The effectiveness of MT-GPRMachine is further validated on the several real-world datasets (detailed prediction performances for all methods are listed in **Table S3**).

In predicting the oxygen content on the plankton dataset (**Fig. 4a**), MT-GPRMachine got 0.1209, 0.1563 and 0.8564 in terms of MAE, RMSE and PCC, respectively (**Fig. 4c-e**), which indicates that MT-GPRMahcine performed much better than the second-best baseline by 65.5% and 66.8% lower in terms of MAE and RMSE, respectively, while 64.8% higher in terms of PCC.

For the prediction of the average temperature on ground ozone level dataset (**Fig. 4b**), we first smoothed the data by the moving average scheme with window length set to 5 due to the drastic fluctuation, then made a prediction based on the smoothed data. Considerable low MAE, RMSE and high PCC were obtained compared to the smoothed data, which are 0.9950, 1.2291 and 0.9401, respectively (**Fig. 4f-h**). All those prediction results show that MT-GPRMachine effectively transformed the high dimensional spatial information into the temporal dynamics of a target variable, and multi-step-ahead predictions were successfully made without any extra information in the future state.

On the prediction of typhoon eye trace (**Fig. 5**), MT-GPRMachine provided pretty good prediction results: 0.0879, 0.1035 and 0.9921 on latitude, 0.135, 0.1645 and 0.9966 on longitude in terms of MAE, RMSE and PCC, respectively. In general, MT-GPRMachine outperformed traditional statistical methods and neural network-based methods. While in many cases, traditional methods failed to make accurate predictions with very high MAE and RMSE and considerably low PCC (even in the rare case that with high PCC, e.g., ARIMA in predicting the longitude of the typhoon eye trace).

As a final test, wireless channel data was collected and applied. After preprocessing (the detailed description can be found in the **SI**), the input training matrix contains 192 variables and 64 features, and its elements are complex numbers. Hence, the real (amplitude) part and imaginary (frequency/phase) part are predicted separately. As shown in **Fig. 6a** and **b**, the dynamical evolutions of the real and imaginary part predicted by MT-GPRMachine are consistent with the real situation, and the corresponding PCCs are 0.843 and 0.844, respectively. In addition, we also consider the presence of missing signals in the wireless channel data, which is very common in practice. For simulating the case with missing values, 10% variables of the training matrix were selected and then 10% values of them were randomly removed. In doing so, the prediction performance of MT-GPRMachine is still competitive, the PCCs of real and imaginary parts are 0.814 and 0.881, respectively (**Fig. 6c and d**), which suggests that MT-GPRMachine is robust with missing values.

## V. CONCLUSION AND DISCUSSION

In many fields of research and industry, it is imperative to develop an effective method for predicting the evolution of target variables from only short-term observed time series. We proposed MT-GPRMachine in this work, a multi-step-ahead prediction method for time series prediction on the basis of the multi-task learning scheme, which makes the multi-step-ahead prediction by simultaneously solving the mappings in STI matrix through multiple linked GPRs. Its effectiveness and robustness were demonstrated on both synthetic and real-world datasets, and compared with other 13 existing methods. The prediction results show that MT-GPRMachine performed superior in terms of accuracy, noise resistance ability and robustness which benefits from the combination of the statistical and dynamical methods, i.e. GPR and STI equation with the multi-task learning scheme. In other words, the major contribution in this paper is that we design a specific multi-task GPR that simultaneously solves the nonlinear STI mappings, and thus provides accurate multi-step-ahead prediction in an efficient and effective manner.

MT-GPRMachine can make accurate predictions with high confidence (low variance) from a short time series, which is an advantage compared to the neural network-based machine learning methods, and is important in real-world applications. Notice that in predicting the average temperature on ground ozone level dataset, we first transformed the original data in a relatively smooth case due to strong fluctuations, which are generally difficult to be trained by GPR-based methods. Even though the prediction results are perfect, MT-GPRMachine still suffers from a few limitations (accurately speaking, it is for all GPR-based methods), e.g. MT-GPRMachine is more suitable for making a prediction of the smooth time-series data. Accordingly, developing an appropriate model/way to combine with GPR for overcoming those kinds of defects is still an open problem for time series prediction, which is also our future topic.

## ACKNOWLEDGMENT

We thank HUAWEI Technologies Co., Ltd. for providing the wireless channel dataset.

## REFERENCES

[1] T. N. Palmer, "Extended-Range Atmospheric Prediction and the Lorenz Model," (in English), *Bulletin of the American Meteorological Society*, vol. 74, no. 1, pp. 49-65, Jan 1993.

[2] D. J. Lockhart and E. A. Winzeler, "Genomics, gene expression and DNA arrays," (in English), *Nature*, vol. 405, no. 6788, pp. 827-836, Jun 15 2000.

[3] H. De Jong, "Modeling and simulation of genetic regulatory systems: A literature review," (in English), *Journal of Computational Biology*, vol. 9, no. 1, pp. 67-103, 2002.

[4] R. R. Stein *et al.*, "Ecological Modeling from Time-Series Inference: Insight into Dynamics and Stability of Intestinal Microbiota," (in English), *Plos Computational Biology*, vol. 9, no. 12, Dec 2013.

[5] M. M. Rienecker *et al.*, "MERRA: NASA's Modern-Era Retrospective Analysis for Research and Applications," (in English), *Journal of Climate*, vol. 24, no. 14, pp. 3624-3648, Jul 2011.

[6] H. S. Cai, X. D. Jia, J. S. Feng, W. Z. Li, Y. M. Hsu, and J. Lee, "Gaussian Process Regression for numerical wind speed prediction enhancement," (in English), *Renewable Energy*, vol. 146, pp. 2112-2123, Feb 2020.

[7] A. Sorjamaa, J. Hao, N. Reyhani, Y. N. Ji, and A. Lendasse, "Methodology for long-term prediction of time series," (in English), *Neurocomputing*, vol. 70, no. 16-18, pp. 2861-2869, Oct 2007.

[8] L. Zhang, W. D. Zhou, P. C. Chang, J. W. Yang, and F. Z. Li, "Iterated time series prediction with multiple support vector regression models," (in English), *Neurocomputing*, vol. 99, pp. 411-422, Jan 1 2013.

[9] Y. C. Ouyang and H. J. Yin, "Multi-Step Time Series Forecasting with an Ensemble of Varied Length Mixture Models," *International Journal of Neural Systems*, vol. 28, no. 4, May 2018.

[10] B. K. Nelson, "Statistical methodology: V. Time series analysis using autoregressive integrated moving average (ARIMA) models," (in English), *Academic Emergency Medicine*, vol. 5, no. 7, pp. 739-744, Jul 1998.

[11] T. Petukhova, D. Ojkic, B. McEwen, R. Deardon, and Z. Poljak, "Assessment of autoregressive integrated moving average (ARIMA), generalized linear autoregressive moving average (GLARMA), and random forest (RF) time series regression models for predicting influenza A virus frequency in swine in Ontario, Canada," (in English), *Plos One*, vol. 13, no. 6, Jun 1 2018.

[12] L. Cui, L. B. Cheng, X. M. Jiang, Z. F. Chen, and Albarka, "Robust estimation and outlier detection based on linear regression model," (in English), *Journal of Intelligent & Fuzzy Systems*, vol. 37, no. 4, pp. 4657-4664, 2019.

[13] R. G. Brown, "Exponential Smoothing for Predicting Demand," (in English), *Operations Research*, vol. 5, no. 1, pp. 145-145, 1957.

[14] Q. T. Tran, L. Hao, and Q. K. Trinh, "A comprehensive research on exponential smoothing methods in modeling and forecasting

cellular traffic," (in English), *Concurrency and Computation-Practice & Experience,* vol. e5602, Dec 5 2019.

[15] Y. Nardi and A. Rinaldo, "Autoregressive process modeling via the Lasso procedure," (in English), *Journal of Multivariate Analysis,* vol. 102, no. 3, pp. 528-549, Mar 2011.

[16] P. Chen, R. Liu, K. Aihara, and L. Chen, "Autoreservoir computing for multistep ahead prediction based on the spatiotemporal information transformation," *Nature Communications,* p. in press, 2020.

[17] C. K. I. Williams, "Prediction with Gaussian processes: From linear regression to linear prediction and beyond," (in English), *Learning in Graphical Models,* vol. 89, pp. 599-621, 1998.

[18] C. K. I Williams and C. E. Rasmussen, "Gaussian processes for regression," (in English), *Advances in Neural Information Processing Systems 8,* vol. 8, pp. 514-520, 1996.

[19] T. H. Liu, H. K. Wei, S. X. Liu, and K. J. Zhang, "Industrial time series forecasting based on improved Gaussian process regression," (in English), *Soft Computing,* Apr 30 2017.

[20] S. Eleftheriadis, T. F. W. Nicholson, M. P. Deisenroth, and J. Hensman, "Identification of Gaussian Process State Space Models," *Proceedings of the 31st International Conference on Neural Information Processing System,* pp. 5315–5325, 2017.

[21] J. A. Han and X. P. Zhang, "Financial Time Series Volatility Analysis Using Gaussian Process State-Space Models," (in English), *2015 Ieee Global Conference on Signal and Information Processing (Globalsip),* pp. 358-362, 2015.

[22] E. Snelson, C. E. Rasmussen, and Z. Ghahramani, "Warped Gaussian Processes," *International Conference on Neural Information Processing Systems,* 2004.

[23] R. Calandra, J. Peters, C. Rasmussen, and M. Deisenroth, "Manifold Gaussian Processes for regression," *2016 International Joint Conference on Neural Networks (IJCNN),* pp. 3338-3345, 2016.

[24] E. V. Bonilla, K. M. A. Chai, and C. K. I. Williams, "Multi-task gaussian process prediction," *Advances in Neural Information Processing Systems 8,* vol. 20, pp. 153-160, 2008.

[25] F. Yousefi, M. T. Smith, and M. Alvarez, "Multi-task learning for aggregated data using Gaussian processes," *Advances in Neural Information Processing Systems 8,* vol. 32, pp. 15076-15086, 2019.

[26] R. Durichen, M. A. F. Pimentel, F. Clifton, A. Schweikard, and D. A. Clifton, "Multi-task Gaussian process models for biomedical applications," *IEEE-EMBS International Conference on Biomedical and Health Informatics (BHI),* pp. 492-495, 2014.

[27] R. Durichen, M. A. F. Pimentel, L. Clifton, A. Schweikard, and D. A. Clifton, "Multitask Gaussian Processes for Multivariate Physiological Time-Series Analysis," (in English), *Ieee Transactions on Biomedical Engineering,* vol. 62, no. 1, pp. 314-322, Jan 2015.

[28] S. Hochreiter and J. Schmidhuber, "Long short-term memory," (in English), *Neural Computation,* vol. 9, no. 8, pp. 1735-1780, Nov 15 1997.

[29] P. R. Vlachas, W. Byeon, Z. Y. Wan, T. P. Sapsis, and P. Koumoutsakos, "Data-driven forecasting of high-dimensional chaotic systems with long short-term memory networks," (in English), *Proceedings of the Royal Society a-Mathematical Physical and Engineering Sciences,* vol. 474, no. 2213, May 2018.

[30] J. Pathak, B. Hunt, M. Girvan, Z. X. Lu, and E. Ott, "Model-Free Prediction of Large Spatiotemporally Chaotic Systems from Data: A Reservoir Computing Approach," (in English), *Physical Review Letters,* vol. 120, no. 2, Jan 12 2018.

[31] A. Haluszczynski and C. Rath, "Good and bad predictions: Assessing and improving the replication of chaotic attractors by means of reservoir computing," (in English), *Chaos,* vol. 29, no. 10, Oct 2019.

[32] N. H. Packard, J. P. Crutchfield, J. D. Farmer, and R. S. Shaw, "Geometry from a Time-Series," (in English), *Physical Review Letters,* vol. 45, no. 9, pp. 712-716, 1980.

[33] E. R. Deyle and G. Sugihara, "Generalized Theorems for Nonlinear State Space Reconstruction," (in English), *Plos One,* vol. 6, no. 3, Mar 31 2011.

[34] T. Sauer, J. A. Yorke, and M. Casdagli, "Embedology," (in English), *Journal of Statistical Physics,* vol. 65, no. 3-4, pp. 579-616, Nov 1991.

[35] F. Takens, "Detecting strange attractors in turbulence," *Lecture notes in mathematics,* vol. 898, pp. 366-381, 1981.

[36] H. Liu *et al.*, "Nonlinear dynamic features and co-predictability of the Georges Bank fish community," (in English), *Marine Ecology Progress Series,* vol. 464, pp. 195-U228, 2012.

[37] H. Ye and G. Sugihara, "Information leverage in interconnected ecosystems: Overcoming the curse of dimensionality," (in English), *Science,* vol. 353, no. 6302, pp. 922-925, Aug 26 2016.

[38] H. F. Ma, S. Y. Leng, K. Aihara, W. Lin, and L. N. Chen, "Randomly distributed embedding making short-term high-dimensional data predictable," (in English), *Proceedings of the National Academy of Sciences of the United States of America,* vol. 115, no. 43, pp. E9994-E10002, Oct 23 2018.

[39] M. S. Masnadi-Shirazi, "Attractor Ranked Radial Basis Function Network: A Nonparametric Forecasting Approach for Chaotic Dynamic Systems," *Scientific Report,* vol. 10, no. 3780, 2020.

[40] C. Chen *et al.*, "Predicting future dynamics from short-term time series by anticipated learning machine," *National Science Review,* Journal vol. 6, 2020.

[41] H. Widom, "Hankel Matrices," (in English), *Transactions of the American Mathematical Society,* vol. 121, no. 1, pp. 1-&, 1966.

[42] J. Berkson, "Bayes' Theorem," (in English), *Annals of Mathematical Statistics,* vol. 1, no. 1, pp. 42-56, Feb 1930.

[43] J. H. Curry, "Generalized Lorentz System," (in English), *Communications in Mathematical Physics,* vol. 60, no. 3, pp. 193-204, 1978.

[44] M. W. Hirsch and S. Smale, "Differential equations, dynamical systems, and linear algebra," *Academic press, New York San Francisco London,* 1974.

[45] A. G. D. Matthews *et al.*, "GPflow: A Gaussian Process Library using TensorFlow," (in English), *Journal of Machine Learning Research,* vol. 18, pp. 1-6, 2017.

[46] F. Pedregosa *et al.*, "Scikit-learn: Machine Learning in Python," (in English), *Journal of Machine Learning Research,* vol. 12, pp. 2825-2830, Oct 2011.

[47] H. Lutkepohl, "Vector Autoregressive and Vector Error Correction Models," *Cambridge University Press,* pp. 86-158, 2004.

[48] H. Lutkepohl, "New introduction to multiple time series analysis," *Springer Science & Business Media, Berlin/Heidelberg,* 2005.

[49] V. Vapnik, "The nature of statistical learning theory," *Springer Science & Business Media, Berlin/Heidelberg,* 2013.

[50] J. D. Farmer and J. J. Sidorowich, "Predicting Chaotic Time-Series," (in English), *Physical Review Letters,* vol. 59, no. 8, pp. 845-848, Aug 24 1987.

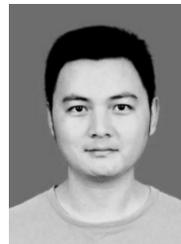

**Peng Tao** received the M.S. and Ph.D. degree from Huazhong University of Science and Technology, Wuhan, China, in 2020. He is currently a postdoc fellow in Hangzhou Institute for Advanced Study, University of Chinese Academy of Sciences, Chinese Academy of Sciences, Hangzhou 310024, China. His current research interests include machine learning, molecular dynamics and computational biology.

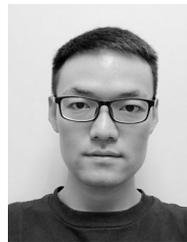

**Xiaohu Hao** received the M.S. and Ph.D. degree from Zhejiang University of Technology, Hangzhou, China, in 2019. He is currently a postdoc fellow in Key Laboratory of Systems Biology, Shanghai Institute of Biochemistry and Cell Biology, Center for Excellence in Molecular Cell Science, Chinese Academy of Sciences, Shanghai 200031 China. His current research interests include machine learning, nonlinear dynamics and computational biology and bioinformatics.

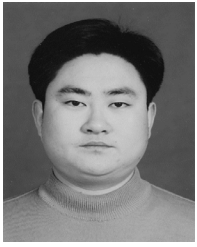

**Jie Cheng** received the M.S. degree from Huazhong University of Science and Technology, in 2011. He is currently an engineer at HUAWEI Technologies Co., Ltd, Shenzhen 518129 China. His research interests include artificial intelligence, brain-inspired computing and machine learning.

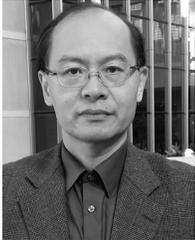

**Luonan Chen** received the B.S. degree from the Huazhong University of Science and Technology, Wuhan, China, in 1984, and the M.S. and Ph.D. degrees from Tohoku University, Sendai, Japan, in 1988 and 1991, respectively. Since 1997, he has been an Associate Professor with Osaka Sangyo University, Osaka, Japan, where he became a Full Professor. Since 2010, he has been a Professor and the Executive Director with the Key Laboratory of Systems Biology, Institute of Biochemistry and Cell Biology, Shanghai Institutes for Biological Sciences, Chinese Academy of Sciences, Shanghai, China. In recent years, he published over 300 journal papers and two monographs in the area of systems' biology. His current research interests include computational systems biology and nonlinear dynamics.